\documentclass{article}

\usepackage{microtype}
\usepackage{graphicx}
\usepackage{subfigure}
\usepackage{enumitem}
\usepackage{booktabs} 

\usepackage{hyperref}

\usepackage[accepted]{icml2023}

\usepackage{amsmath}
\usepackage{amssymb}
\usepackage{mathtools}
\usepackage{amsthm}

\usepackage[capitalize,noabbrev]{cleveref}

\theoremstyle{plain}

\theoremstyle{definition}

\theoremstyle{remark}

\newcommand{\bigO}[1]{$\mathcal{O}(#1)$}

\usepackage[textsize=tiny]{todonotes}

\def \reusabletitle {SpeedLimit: Neural 
Architecture Search for Quantized Transformer Models}

\icmltitlerunning{\reusabletitle{}}

\begin{document}

\twocolumn[
\icmltitle{\reusabletitle{}}



\icmlsetsymbol{equal}{*}

\begin{icmlauthorlist}
\icmlauthor{Yuji Chai}{equal,harvard,stochastic}
\icmlauthor{Luke Bailey}{equal,harvard}
\icmlauthor{Yunho Jin}{harvard}
\icmlauthor{Matthew Karle}{harvard}
\icmlauthor{Glenn G. Ko}{harvard,stochastic}\\
\icmlauthor{David Brooks}{harvard}
\icmlauthor{Gu-Yeon Wei}{harvard}
\icmlauthor{H. T. Kung}{harvard}
\end{icmlauthorlist}

\icmlaffiliation{harvard}{Harvard University}
\icmlaffiliation{stochastic}{Stochastic Inc}

\icmlcorrespondingauthor{Luke Bailey}{lukebailey@college.harvard.edu}

\icmlkeywords{
Machine Learning, 
ICML,
Neural Architechture Search,
Quantization,
Transformers,
LLM
}

\vskip 0.3in
]



\printAffiliationsAndNotice{\icmlEqualContribution} 


\begin{abstract}
While research in the field of transformer models has primarily focused on enhancing performance metrics such as accuracy
and perplexity, practical applications in industry often necessitate a rigorous consideration of inference latency constraints. Addressing this challenge, we introduce  \emph{SpeedLimit}, a novel Neural Architecture Search (NAS) technique that optimizes accuracy whilst adhering to an upper-bound latency constraint. Our method incorporates 8-bit integer quantization in the search process to outperform the current state-of-the-art technique. Our results underline the feasibility and efficacy of seeking an optimal balance between performance and latency, providing new avenues for deploying state-of-the-art transformer models in latency-sensitive environments.
\end{abstract}

\section{Introduction}
\label{section:Introduction}

Transformer models 
are incredibly capable across diverse domains,
most notably natural language processing \cite{GPT3, GPT2, BERT, ROBERTA}
and computer vision \cite{IMAGE-RECOG-TRANSFORMERS, IMAGE-TRANSFORMER, VISION-TRANSFORMERS}. 
In recent years, transformer capability has been 
driven in large parts by expanding model sizes
\cite{SCALING}. With state-of-the-art (SOTA) models now 
exceeding hundreds of gigabytes in size \cite{PALM, MEGATRON-TURING},
it has 
become incredibly costly to deploy and maintain 
services that rely on them. Additionally, the models
are so large that
low-latency deployments have become impossible without paying
for the most advanced hardware. 
In many 
application areas, upper bound latency restrictions are required, 
for example chatbot services or real
time object recognition. These applications occur
across many diverse settings, from large cloud
computing clusters to small edge devices. 
In
these scenarios, the latency requirements are fixed
and machine learning engineers must find the most
performant
model that meets said requirement. To solve this
problem we propose \emph{SpeedLimit}, a
neural architecture search (NAS) technique 
for finding optimal transformer models
with fixed upper bound latency
constraints.


In this paper we focus on BERT \cite{BERT} as our motivating transformer 
architecture. The current SOTA NAS process to find BERT models that satisfy a latency requirement, called AutoTinyBERT \cite{autotinybert}, focuses solely on models with float32 parameters. Seen as the goal of the algorithm is to find the most accurate models at a certain latency target, it seems naive to not incorporate 8-bit integer (int8) quantization into this process in light of the inference speedup associated with quantized models on increasingly common commodity hardware. 
Additionally, as we will show, naively quantizing the float32 model that other NAS techniques create does not guarantee the best int8 model. We thus aim to fix this problem by incorporating quantization into the NAS process itself. This means we search directly for the best int8 model at a given latency constraint.

\begin{figure}[]
\centerline{\includegraphics[width=0.45\textwidth]{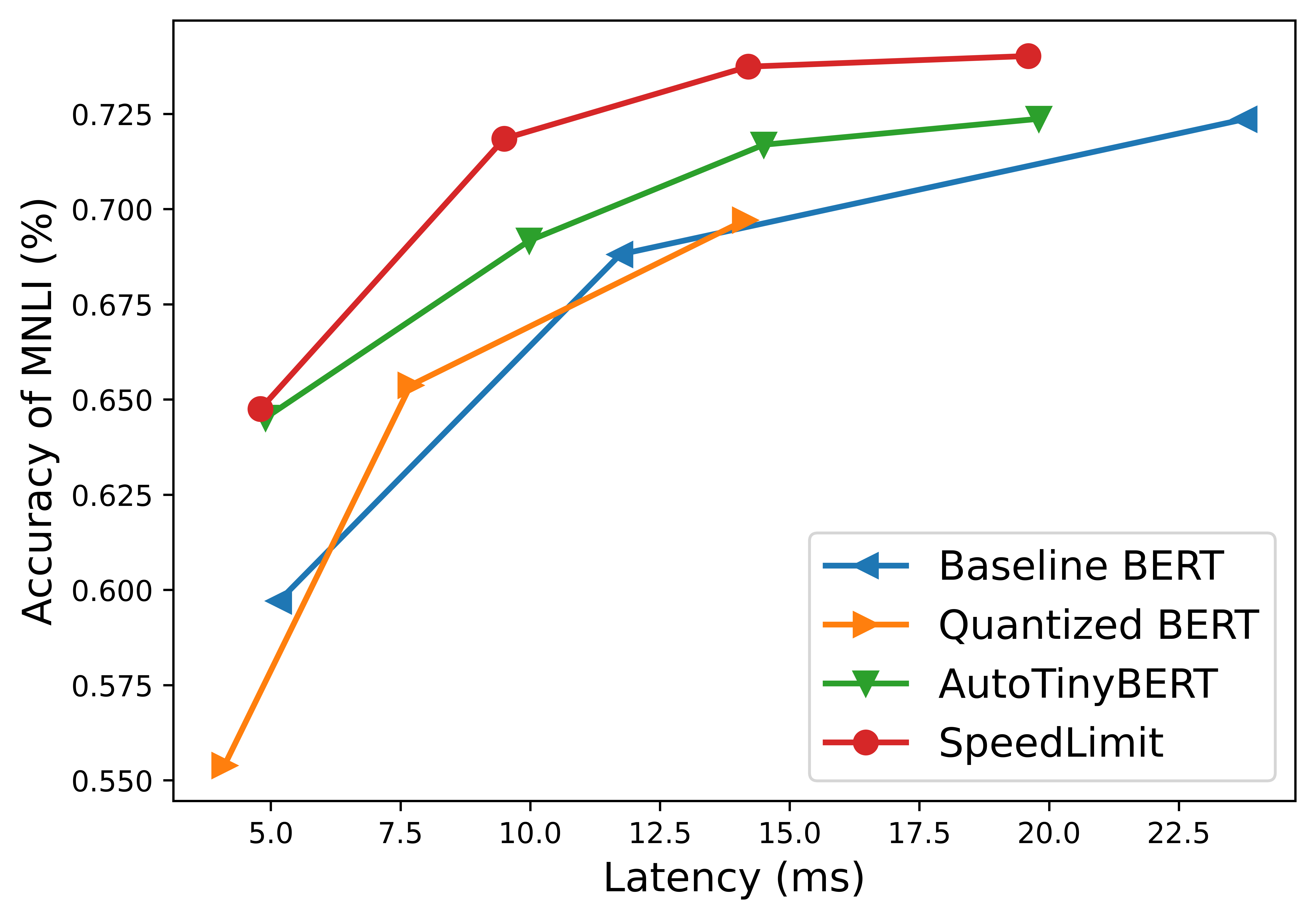}}
\caption{Latency vs. MNLI task accuracy.
\emph{SpeedLimit} outperforms AutoTinyBERT
\cite{autotinybert}, full precision BERT (baseline),
and int8 quantized BERT (Quantized) across all latency 
requirements tested.}
\label{figure:mnli}
\vspace{-5mm}
\end{figure}

\begin{figure*}[t] 
\centerline{\includegraphics[width=0.8\textwidth]
{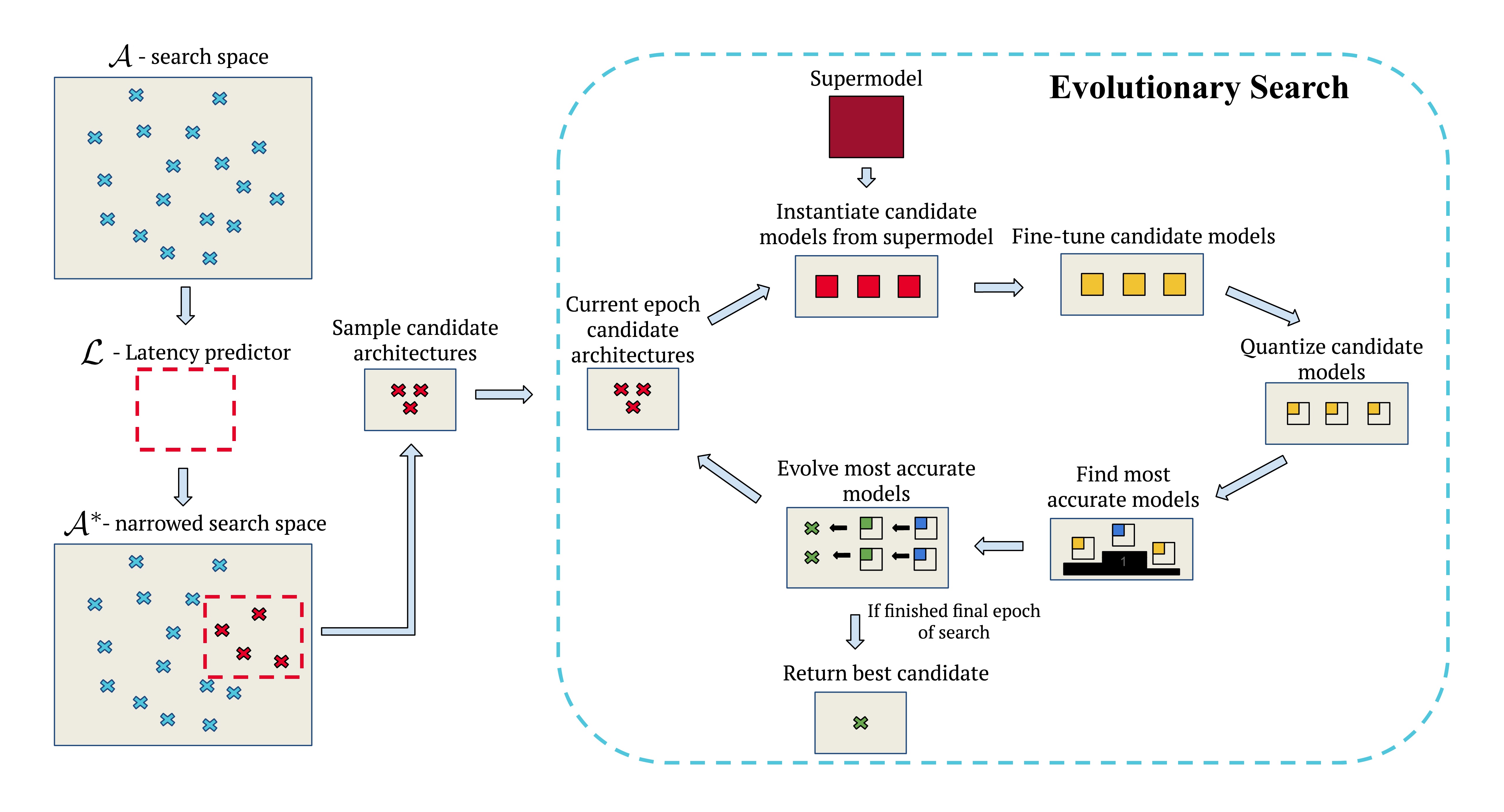}} 
\caption{\emph{SpeedLimit} pipeline.} 
\label{figure:pipeline}
\end{figure*}

\emph{SpeedLimit} applies a two-stage NAS technique to find such an optimal int8 quantized architecture \cite{cai2020once}. Our approach involves training a Supermodel via knowledge distillation, and then employing an evolutionary search to find the best quantized subnetworks of this Supermodel, called candidates. Two-stage NAS streamlines the evaluation process by eliminating the need for extensive training from scratch, which is particularly vital for large SOTA transformers. Candidates are assessed based on their latency and accuracy, with the top-performing one chosen as the final model.
\emph{SpeedLimit} is able to find models that outperform those
found by AutoTinyBERT
and a baseline of default BERT models (both full precision and 
int8 quantized) in terms of accuracy and latency.



\section{Background and related work}
\label{section:background}


Neural architecture search (NAS) is a technique for 
automatically finding
optimal deep neural network architectures \cite{NAS-SURVEY}. 
NAS techniques define 
some search space of 
architectures, and a 
search strategy over this 
space. Common search 
algorithms include
evolutionary search, 
reinforcement learning, Bayesian optimization, and gradient
based methods \cite{NAS-SURVEY}. 
A search algorithm works in tandem with some 
performance estimation strategy
that the algorithm uses
to query for the runtime performance 
of each architecture
during the search.
NAS techniques can be
broadly classified into two
categories according
to their performance estimation 
strategy: one-stage and two-stage. 

In one-stage NAS, the performance
of an architecture is calculated
by simply instantiating it, 
training it from scratch, and 
evaluating the resulting model.
In two-stage NAS, before performing 
the search, a Supermodel (that is larger 
than any candidate architecture
in the search space) is
trained. Then during the search 
process, the performance of 
architectures is estimated by
instantiating them 
with weights extracted from the
Supermodel and either evaluating 
the resulting
models immediately or after
a small number of fine-tuning 
steps  \cite{cai2020once}.
Because
of the reduction in training
candidate architectures, two-stage
NAS accelerates 
the search process significantly, especially for large
models. AutoTinyBERT \cite{autotinybert} was able to
use two-stage NAS to produce models with reduced inference
latency without loss in 
performance on the
GLUE benchmark \cite{autotinybert}
compared to the SOTA search-based method and distillation-based methods \cite{NAS-BERT, DISTILBERT,TINYBERT, MINILM, MOBILEBERT}.
AutoTinyBERT specifically uses NAS to create models that satisfy certain latency constraints, whilst maximizing model accuracy, as opposed to targeting accuracy directly.

Quantization is another technique aimed at model compression by converting high-precision floating-point parameter values to lower-precision data types, which in turn reduces memory footprint and can speed up model inference if the
target hardware supports faster computation on
said quantized data types. Specifically, accelerated 
int8 quantized inference is now supported by server-grade
CPUs and GPUs, allowing such models to include more
parameters compared to float32 models under the same latency constraint. This opens up a new direction for optimization by trading parameters' precision for larger parameter counts, which inspires this work. Prior works such as I-BERT \cite{ibert} and Q8BERT \cite{q8bert} have successfully applied quantization to BERT, maintaining high test accuracy while achieving model compression and inference speedup.

\begin{figure*}[ht!]
     \centering
     \includegraphics[width=\textwidth]{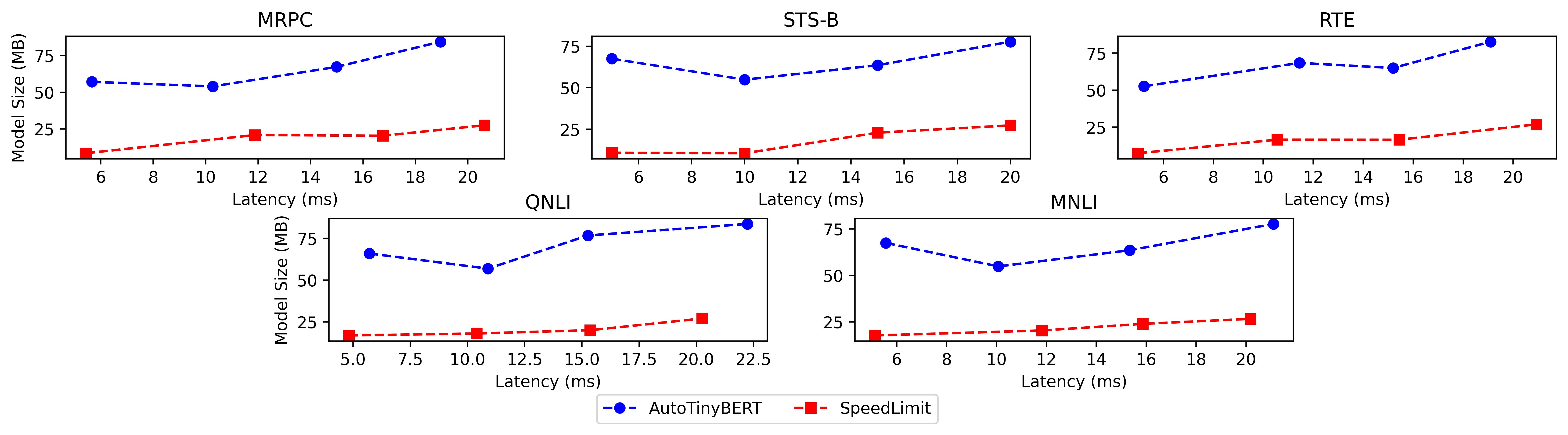}
     \caption{Latency and sizes of models found
     by \emph{SpeedLimit} and AutoTinyBERT.}
     \label{fig:model_size}
     \vspace{-3mm}
\end{figure*}

\begin{figure}[]
\centerline{\includegraphics[width=0.45\textwidth]{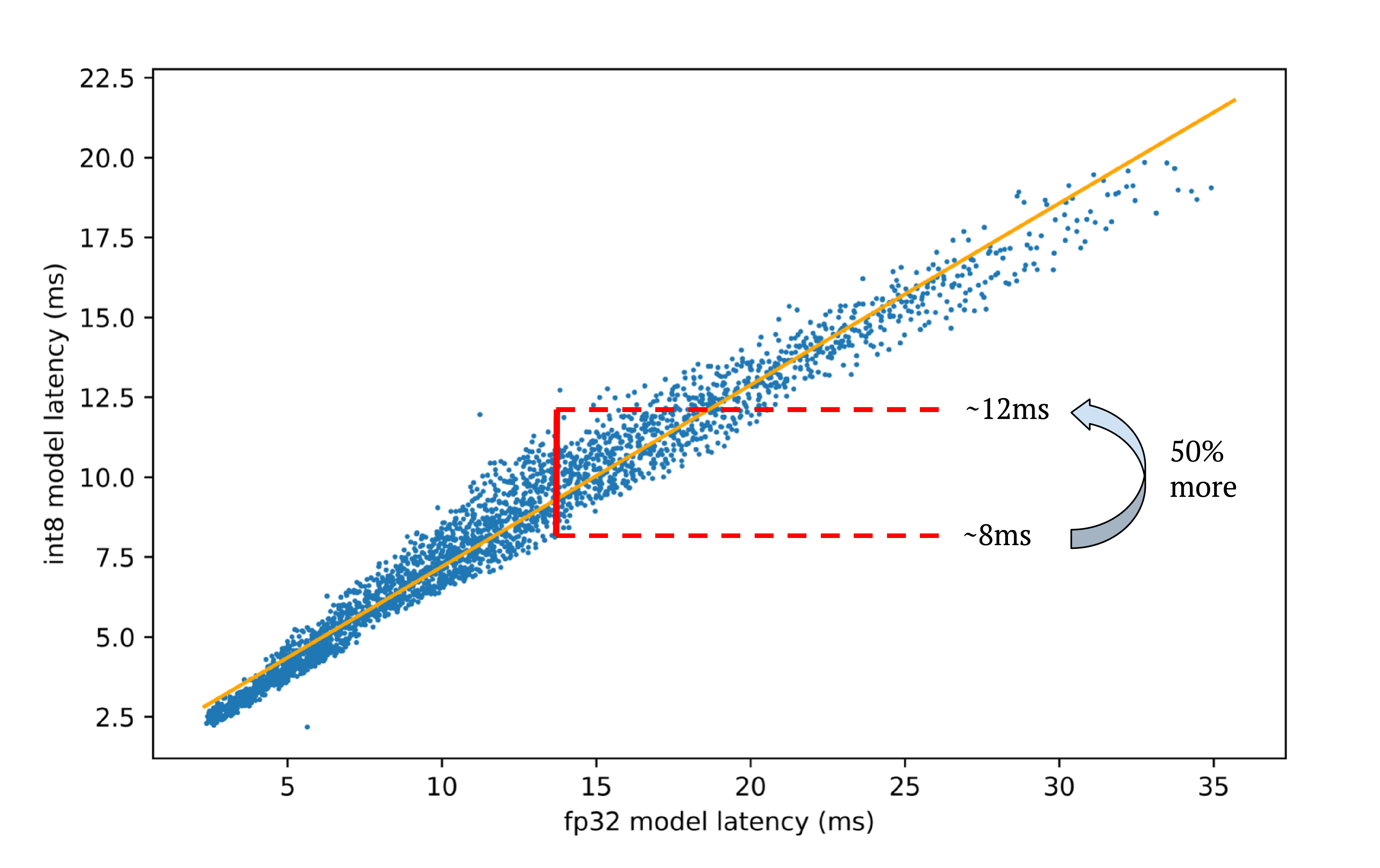}}
\caption{Comparison of various models' int8 version and float32 version latencies. The same float32 and int8 model configuration can result in very different corresponding latency values.}
\label{figure:latency_compare}
\vspace{-4mm}
\end{figure}

Finally, the 2018 JASQ method \cite{jasq} combined 
quantization with NAS, and helped to motivate our
contributions, but is ultimately a very different endeavor.
JASQ searches both architectures and quantization policies to produce an optimal CNN with quantization precision potentially mixed across layers. 
Unlike our method, 
they target much smaller 
models and only in the vision and not 
language domain. Accordingly, they employ 
one-stage as opposed to two-stage NAS. 
Additionally, their method targets 
a certain model size constraint as 
opposed to latency, hence their use of
mixed precision quantization. We solely
target int8 quantization as it has more
widely supported computational speedup 
compared to other quantized datatypes.



\section{Methodology}
\label{section:Methodology}

We propose \emph{SpeedLimit}, a method to automatically search for 
an optimal quantized Transformer architecture given a latency
constraint. By optimal, we mean
maximizing the performance of the model by
some metric on a downstream fine-tuning task
whilst still hitting an inference latency constraint. We
use BERT as an example Transformer, however 
our techniques can easily be applied to other 
Transformer based models. 


Figure \ref{figure:pipeline} shows the full
\emph{SpeedLimit} pipeline.
Firstly, given some latency constraint, we use a latency predictor 
to narrow the search space of all
possible architectures to just that of int8 architectures that will meet 
the inputted latency constraint. We call the resulting search space
the narrowed search space. 
The use of the latency predictor greatly accelerates the 
search space narrowing process, as it removes the need to 
profile the performance of all the model
architectures present in the original search space.
Next,
we apply two-stage NAS to the narrowed search space,
using iterative rounds of evolutionary search to find the
optimal BERT configuration. As we use
two-stage NAS, before conducting the 
search we train a large Supermodel. 
At each round of the search, 
candidate models are
extracted from this Supermodel, fine-tuned 
for a small
number of epochs, 
quantized and evaluated. For quantization,
we use the PyTorch 
implementation of dynamic quantization, however
this can easily be replaces with other 
post-training quantization techniques 
such as GPTQ \cite{frantar2022gptq}.
The best models are evolved 
and make up the candidates for the following round.
After a user inputted number of rounds, the best performing
current candidate is returned.

\begin{figure*}[h!]
     \centering
     \includegraphics[width=\textwidth]{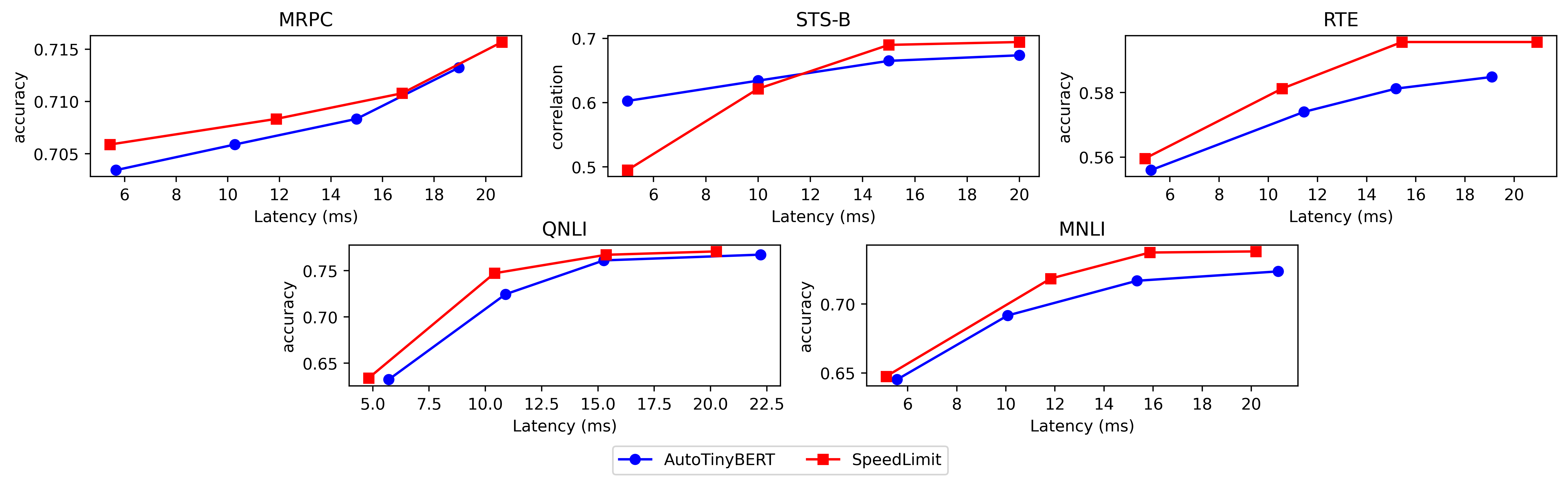}
     \caption{Accuracy and latency values of
     models found by \emph{SpeedLimit} and AutoTinyBERT.}
     \label{fig:acc_vs_lat}
     \vspace{-2mm}
\end{figure*}

\textbf{Search space:} BERT contains many architecture 
hyper-parameters that can be changed.
In this work, we restrict our search space
to only contain architectures that have homogeneous 
encoder blocks (the architecture of each encoder
block is identical). 
Let $e$ be the number of encoder 
blocks present
in the model and $d$ be the number of different 
individual encoder block architectures. Using
homogeneous encoder blocks reduces our search 
space from polynomial order $e$, \bigO{d^e}, to 
simply linear \bigO{d}. This reduction in 
search space size is essential because even
two-stage NAS is very computationally intensive.
We consider varying $h$, the dimensionality of encoder, attention head, and
pooler layers, $f$, the size of the dense feed forward layers,
and $e$. With these
hyperparameters, each model architecture we 
consider can be encoded in a 
three element tuple ($e$, $h$, $f$). Let $\mathcal{A}$ be the 
set of all architectures we consider. Naturally we must 
bound the values of $e$, $h$ and $f$ to ensure
that $|\mathcal{A}|$ is small enough to allow searches in reasonable times. 
Table \ref{tab:dataset_table} in the Appendix
presents the architectures we used during testing.  


Given a user-inputted latency constraint $l$, we aim to 
search only the architectures that correspond to int8
quantized models meeting this constraint. We define
$\mathcal{A}^*$ to be the set of such models, 
which we call the narrowed search space.
Ideally, we would 
instantiate every
architecture in $\mathcal{A}$, quantize 
them, find their latencies, 
and only places architectures in $\mathcal{A}^*$
that meet $l$. Unfortunately, this 
is incredibly time-consuming because 
instantiating and quantizing large transformer
models takes a non-trivial amount of time,
and $|\mathcal{A}|$ is large. Instead, 
we take a small subset of $\mathcal{A}$,
find the latencies of the quantized models
corresponding to these architectures, 
train a small latency predictor $\mathcal{L}$ on
this data, and use this latency predictor 
to find $\mathcal{A}^*$. Overall, this process
takes far less time than checking the latencies
of all architectures in $\mathcal{A}$.
More formally, after training $\mathcal{L}$,
we find the narrowed search space as 
$\mathcal{A}^* = \{a \in \mathcal{A} : \mathcal{L}(a) < l \}$. 
For more details on the exact training 
procedure for $\mathcal{L}$ used
in our experiments, see section 
\ref{sec:exp_details} of the Appendix.

\textbf{Supermodel:} During evolutionary search, candidate models are selected 
and instantiated using weights extracted from the Supermodel. For
this reason, we need the Supermodel to have an architecture that 
is larger than any of those contained within $\mathcal{A}$.
Let $A_s$ be the architecture of the Supermodel and $\mathcal{E}$,
$\mathcal{H}$, $\mathcal{F}$ be the sets of $e, h$ and $f$ values
present in $\mathcal{A}$. 
Thus, we have:
$A_s = (\text{max}(\mathcal{E}), \text{max}(\mathcal{H}), 
\text{max}(\mathcal{F}))$. We instantiate the Supermodel with architecture $A_s$ and use 
the modified training algorithm proposed in 
\cite{autotinybert} (that 
encourages the model to be more amenable to weight sharing) to train it. See section \ref{sec:super_model_training} 
of the Appendix for more information on the weight 
extraction procedure and Supermodel training algorithm.




\textbf{Conducting evolutionary search:} We adapt the evolutionary search process from AutoTinyBERT
(our algorithm is outlined in section \ref{sec:search_proc} of the Appendix).
The search takes as input 
the narrowed search space $\mathcal{A}^*$.  
For a user inputted number of rounds,
the algorithm extracts the current candidate
architectures from the Supermodel, fine-tunes them,
quantizes them, and evaluates their performance. 
The best performing architectures are evolved
and used as the starting candidates for the next
iteration of search (with the candidates in the first
round of search being chosen randomly from 
$\mathcal{A}^*$). Evolving involves either
mutating the candidate by randomly perturbing 
$(e, h, f)$ values (ensuring the result still 
adheres to the latency constraint) with probability 
$p_m$ or sampling an entirely new candidate with probability 
$1-p_m$. Thus, $p_m$ is a hyperparameter that
controls the rate of exploration of the search algorithm.



\section{Results and Discussion}
\label{section:results}

\textbf{Latency results for BERT models:} 
To motivate why searching for int8 quantized models can outperform 
searching for full precision models, we benchmarked various 
quantized and non-quantized BERT models whose architectures 
were drawn from $\mathcal{A}$ as defined in Table 
\ref{tab:dataset_table} of the Appendix. 
For experiments, we used
an Intel Ice Lake Xeon Platinum 8358 CPU @ 2.60GHz. This CPU was
chosen intentionally as it comes enabled with the AVX-VNNI extension that
allow for accelerated int8 operations. To benchmark the models, we
simply drew a set of architectures and instantiated full floating point
models of each. We recorded the 
averaged latency of every model over 4 single sentence inputs,
each containing 128 tokens (after one untimed run to allow for model warm-up). We then used PyTorch's dynamic quantization library to quantize the models to int8 and re-ran the average latency test. 
Figure \ref{figure:latency_compare} shows the collected
data. Every point represents a
certain architecture for a BERT model. 
We fitted a linear regression to the data to quantify 
the relationship
between int8 and float32 inference times, finding
$latency_{float32} = 1.75 \times latency_{int8}
- 2.65$.

The regression analysis illustrates a significant reduction in model latency when using int8 as compared to float32, indicating the potential for employing a more parameter-dense, quantized model within the same latency constraint. Interestingly,  Figure \ref{figure:latency_compare} shows two distinct models, identical in float32 latency, have a roughly 50\% discrepancy in int8 latency. This phenomenon, pervasive across our dataset, underscores the necessity of an independent, quantization-aware search for int8 models. Relying solely on a float32 model search followed by quantization could result in suboptimal int8 models due to the observed variance in int8 latencies for models with identical float32 latencies.


 
\textbf{Optimal Configurations for int8 BERT Models:} Using the method described in section \ref{section:Methodology} 
we conducted searches for four different latency targets. 
We used $\mathcal{A}$ and 
$\mathcal{L}$ as described 
in section \ref{sec:exp_details} of the 
Appendix. 
For each latency target experiment,
we conducted four rounds of evolutionary search. 
The final search result is shown in 
Figure \ref{fig:acc_vs_lat}. We see our method outperforms 
AutoTinyBERT for all
latency targets bar two for the STS-B dataset. In
the best case, we see up to a 
 2.7\% accuracy gain on the MNLI dataset. 
More generally, a better
search strategy should give a curve closer to the upper left corner, thus being closer to the
Pareto frontier. We see that, across almost all
tasks, \emph{SpeedLimit} beats AutoTinyBERT by this
qualitative metric.
 We also note \cite{autotinybert} reports 
 AutoTinyBERT outperforms standard BERT models with 
 corresponding latencies. We test this specifically using the 
MNLI dataset in
Figure \ref{figure:mnli}. More precisely, we compare
the latency and accuracy of models found
by \emph{SpeedLimit} against 
a set of default BERT architectures fine-tuned using
full float32 precision (Baseline BERT). We also compare 
against these baseline models quantized to int8 
using PyTorch dynamic quantization (Quantized BERT).
As expected (seen as we outperform AutoTinyBERT),
we outperform both of these BERT baselines.

With quantization, we allow the model to have more parameters
while remaining within the latency range restriction.
Although the reduced precision for all the parameters
changing from float32 to int8 will reduce the model's
performance to some extent, this loss is overpowered by the
accuracy gain of additional parameters. We also 
recorded the size of the models that are 
presented in Figure \ref{fig:model_size}. We 
see that in addition to better performance, the
memory footprint of models found by \emph{SpeedLimit}
are on average $3.4\times$ smaller than those found 
by AutoTinyBERT. This reduction in size 
is critical for many applications, especially 
when deploying models such as BERT on
edge devices with low memory resources
\cite{tinymlsurvey}.

\section{Conclusion}

We present \emph{SpeedLimit}, a novel method that combines two-stage NAS and quantization to find latency constrained BERT models. We are able to outperform the current SOTA
method in terms of accuracy, latency, and
memory footprint of outputted models. Our method serves as a key step forward in 
allowing practitioners to deploy 
performant models in strict latency
constrained environments.


\bibliography{paper}
\bibliographystyle{icml2023}

\newpage
\appendix
\onecolumn

\section{Super Model}
\label{sec:super_model_training}

We use the same Supermodel training and
submodel weight extraction technique as 
\cite{autotinybert}. We summarize the training 
scheme here.
We divide each training input batch into
$n$ sub-batches and distribute
them onto $n$ threads. Then for $m$ steps, we sample $n$ sub models from the
Supermodel and distribute them onto the $n$ threads and calculate the
gradient update to be taken. After these $m$ steps have completed,
we average all the gradient updates across the $n$ threads and use this
average gradient to update the model weights. 

For the loss, we use pure knowledge distillation from a BERT-base-uncased
teacher model conducting masked language modelling on the BookCorpus 
\cite{zhu2015aligning} dataset. We use the same hyperparameters
as \cite{autotinybert} (peak learning rate of 1e-5, warm-up rate
of 0.1, $n=16$ and $m=3$) except we use a smaller batch size
of 12, a maximum sequence length of 512, and train for 4 epochs.

\section{Search Procedure}
\label{sec:search_proc}

Algorithm \ref{alg:cap} summarizes the 
evolutionary search algorithm used in 
\emph{SpeedLimit}.

\begin{algorithm}[ht!]
\caption{Evolutionary Search}\label{alg:cap}
\begin{algorithmic}
\STATE {\bfseries Input:} $T$, the number of
generations of evolutionary search, $S$ the number of candidates 
to consider at generation, $p_m$ the mutation probability, $\mathcal{A}^*$
\STATE $\mathcal{G}_1 \gets \mathcal{A}^*$
\FOR{$t=1,2,\dots,T$}
    \STATE $\mathcal{G}_t \gets \{\}$
    \WHILE{$|\mathcal{G}_t| < S$}
        \STATE $\alpha_\text{old} \gets$ a sample (without replacement) from $\mathcal{G}_{t-1}$.
        \STATE $\alpha_\text{quant} \gets$ an 8-bit quantized version of $\alpha_\text{old}$.
        \STATE $p \gets$ a uniform random number from 0 to 1.
        \IF{$p < p_m$}
             \STATE $\alpha_\text{new} \gets$ a mutation of $\alpha_\text{quant}$.
        \ELSE
             \STATE $\alpha_\text{new} \gets$ a random sample from $\mathcal{A}$.
        \ENDIF
        \STATE Append $\alpha_\text{new}$ to $\mathcal{G}_t$
    \ENDWHILE
\ENDFOR
\STATE $\mathcal{M} \gets$ the set of models with architectures from $\mathcal{G}_T$ and weights from the Supermodel.
\STATE $\mathcal{M}_\text{quant} \gets \{\}$
\FOR{$m \in \mathcal{M}$} 
    \STATE Append the quantized version of $m$ to $\mathcal{M}_\text{quant}$
\ENDFOR
\STATE $\alpha_\text{opt} \gets$ the architecture of model with the best accuracy on the target task from $\mathcal{M}_\text{quant}$.
\STATE \textbf{return} $\alpha_\text{opt}$
\end{algorithmic}
\end{algorithm}

\section{Experimental Details}
\label{sec:exp_details}

For our experimental results presented in 
section \ref{section:results}, we 
applied constraints on the $(e, h, f)$ 
values that 
we included in the search space of 
architectures $\mathcal{A}$. This was 
done to ensure that narrowing the
search space to $\mathcal{A}^*$,
set of architectures that met the inputted
latency constraint, could be done in reasonable 
time. The values we selected to be in 
$\mathcal{A}$ are shown in table \ref{tab:dataset_table}.

For the latency predictor $\mathcal{L}$, we used a simple multi layer perceptron with 3 hidden layers and 2000
nodes per layer. We trained
using an Adam optimizer 
with a learning rate 
of $1e-5$, $\beta_1 = 0.9$,
$\beta_2 = 0.99$, 
$\theta_0 = 1e-08$ and
a mean squared error loss. We conducted
a grid search to find the 
optimal learning rate, 
and used the default values
from the PyTorch implementation 
of the Adam optimizer for
$\beta_1, \beta_2$ and $\theta_0$.
We trained the latency predictor
for 5000 steps, each of 
which used a
batch of 128 random architectures from 
$\mathcal{A}$ (as defined in 
table \ref{tab:dataset_table}) labeled
with quantized latencies. To find the
latencies for this training 
data, we instantiated the 
given architecture, quantized 
it, and recorded the latency
across 5 forward passes with 
a single input containing 128 tokens. 
We excluded the first latency value as
they were usually significantly larger
than subsequent values due to model
warm-up (caching of model parameters),
and used the average of the remaining 4
latency values as the latency label.
Our implementation of
$\mathcal{L}$ was able to achieve a 3.28\% mean average
percentage error on a held out testing
set.

\begin{table}[H]
\centering
\caption{Hyperparameter values in search space}
\begin{tabular}{ |c|c| } 
 \hline
 Hyperparameter & Value present in $\mathcal{A}$\\
 \hline
 \hline
 $e$ & $[1,2,3,4,5]$  \\ 
 \hline
  $h$ & $[120, 132, \dots, 12k, \dots, 516, 528]$ \\ 
  \hline
 $f$ & $[128, 140, \dots, 12k, \dots, 1004, 1016]$  \\ 
 \hline
\end{tabular}
\label{tab:dataset_table}
\vspace{-2mm}
\end{table}

\end{document}